\documentclass[12pt]{article}

\usepackage{graphicx}
\usepackage[super]{cite}
\usepackage[letterpaper,margin=1in]{geometry}
\usepackage{amsmath}
\usepackage{hyperref}
\usepackage{caption}
\date{}

\usepackage{url}

\title{
Environmental resilience via morphological diversity within machines
}

\author{
	Alice~Hein$^{}$,
	Josh~Bongard$^{\ast}$ \\
	\small
    Department of Computer Science, 
    University of Vermont, Burlington VT \& 05405, USA.\and
	\small$^\ast$Corresponding author. Email: jbongard@uvm.edu
}

\begin{document} 


\maketitle

\begin{abstract} 
Organisms contain diverse, sensorimotor parts across size scales and rapidly adapt to new environments, while machines contain only inert materials at smaller scales and struggle with surprise. We hypothesize that this agents-within-agents quality of organisms may aid their resilience: increasing experiences with internal physical adversity may pre-train organisms and machines to handle external adversity, such as encounters with new environments. Not only has this hypothesis not yet been articulated, mechanisms enabling this phenomenon have yet to be proposed. Here we show a mechanism by which this can occur: we found that physical connectors, in learning to restore behavior to previously independent, morphologically diverse agents they disrupted by tethering them together, trigger and tame sufficiently diverse disruptions that later encounters with new environments trigger disruptions that fall within this manageable range, enabling the collective to continue behaving properly without any additional learning or adaptation. Further, we found that building collectives from more agents, or more diverse agents, further increases the collective’s resilience to new environments. This suggests that not just taming but intentionally creating internal physical adversity may indeed prepare organisms for external adversity, and could do so for machines, if they were built from smaller machines.
\end{abstract}

\section{Introduction.}

Life, when confronted with unfamiliar situations, often recovers function rapidly, sometimes without the need for extensive learning or adaptation: insects adapt to fluctuations in the availability and quality of nutritional resources \cite{leonhardt2020mechanisms}, bacteria reliably navigate chemical gradients across orders of magnitude in concentration, and embryos establish stable body patterns despite wide molecular variation \cite{kitano2004biological}. In contrast, modern engineered systems, from billion-parameter AI models to robot swarms, still  struggle with novelty: faced with unfamiliar inputs, they can fail to generalize \cite{DBLP:conf/naacl/WuQRA0WKAK24}, lose reliable confidence \cite{gawlikowski2023survey}, or display brittle behavior~\cite{Bjerknes2013,heaven2019deep}. Despite their real and growing successes, these systems highlight a persistent gap: a key mechanism that supports the flexibility of living systems seems to still be missing from many autonomous technologies.

One candidate principle is that life, at every scale, consists of diverse collectives~\cite{tajfel_human_2081, ariel2015order}. Biological systems are often modeled as if their members were identical, but when looked at closely this is rarely the case \cite{tessone2006diversity,altschuler2010cellular,muraille2014generation,szwaykowska2015collective}. Even populations of bacteria that all descend from a single cell will not behave in exactly the same way: some divide rapidly while others slow down, stop dividing, or enter distinct physiological states \cite{balaban2004bacterial,kearns2005cell}. Neurons in the same brain region differ in their shapes, electrical activity, and connections \cite{peng2021morphological}. Within animal groups, individuals may vary in size, energy reserves, or in consistent behavior patterns such as boldness or caution \cite{Ariel2022}. Such differences arise from many sources, including developmental trajectory, random molecular events, age, health, nutrition, and experience \cite{knebel2021collective,Ariel2022}. What is often dismissed as noise is in fact a pervasive feature of living systems: each component is, in some meaningful way, distinct from its peers.

Studies across biology have shown that diversity within groups brings important advantages \cite{roger_gen_div,10.1371/journal.pone.0013382, tessone2006diversity, altschuler2010cellular,Ariel2022,jolles2020role}. In bacteria, differences among species, forms, or motility often make entire colonies more robust—they are better able to move, survive antibiotics, and cope with changing environments \cite{balaban2004bacterial,benisty2015antibiotic,finkelshtein2015bacterial,zuo2020dynamic,bottery2021ecology,peled2021heterogeneous,Ariel2022}. Colonies of social insects like ants or bees perform better when workers differ in their roles and responsiveness, which improves foraging and defense \cite{jeanson2014interindividual,o2021functional}. 

\noindent In animal groups such as shoals of fish or flocks of birds, a blend of bold, cautious, fast, and slow individuals helps the group stay cohesive, evade predators, and find food \cite{hemelrijk2005density,dyer2009shoal,pettit2015speed,jolles2017consistent,del2018importance}. And in humans and component-based simulations, teams with a wider range of perspectives and problem-solving strategies have been shown to outperform more uniform groups \cite{hong2004groups,krause2011swarm,licalzi2012power,marcolino2014give,aminpour2021diversity,karamched2020heterogeneity,barrera2024promoting}. 

While diversity is important for collective intelligence and adaptability, research across biology, psychology, and engineering shows that diversity alone is not enough. Mechanisms for integrating and coordinating individual contributions are needed to realize these benefits \cite{koriat2012two,keuschnigg2017crowd,choudhary2025human}. Simply increasing the variation in a collective yields more potentially productive, but also more potentially counterproductive behavioral responses to a given situation, which can impede consensus, propagate errors, or trap collectives in suboptimal patterns~\cite{giraldeau2002potential,tessone2006diversity,koriat2012two,prasetyo2018best,toyokawa2019social}. Without the right structures in place, heterogeneity can just as easily hinder group performance. 

The way collectives harness diversity is deeply tied to how their members connect and interact. Across scales, nature assembles groups using a wide array of connective architectures: some are chemical, like the selective signals and adhesion molecules that synchronize migrating cells \cite{lecaudey2006organizing,shim2021overriding} or the quorum-sensing cues coordinating gene expression in bacterial colonies \cite{rutherford2012bacterial}; others are mechanical, such as the modular links enabling motor control \cite{bottinelli2000human,mclean2015peeling,purslow2020structure,hug2023common} or physical attachments in bacterial biofilms \cite{maier2021physical} and animal aggregations; still others are informational or social, seen in the way animals communicate, share knowledge, or coordinate roles within a group \cite{couzin2005effective,gelblum2015ant,fonio2016locally,berdahl2018collective,kao2019modular,almaatouq2020adaptive,galesic2023beyond}. 

\begin{figure*}[!t]
\centerline{\includegraphics[width=1\linewidth]{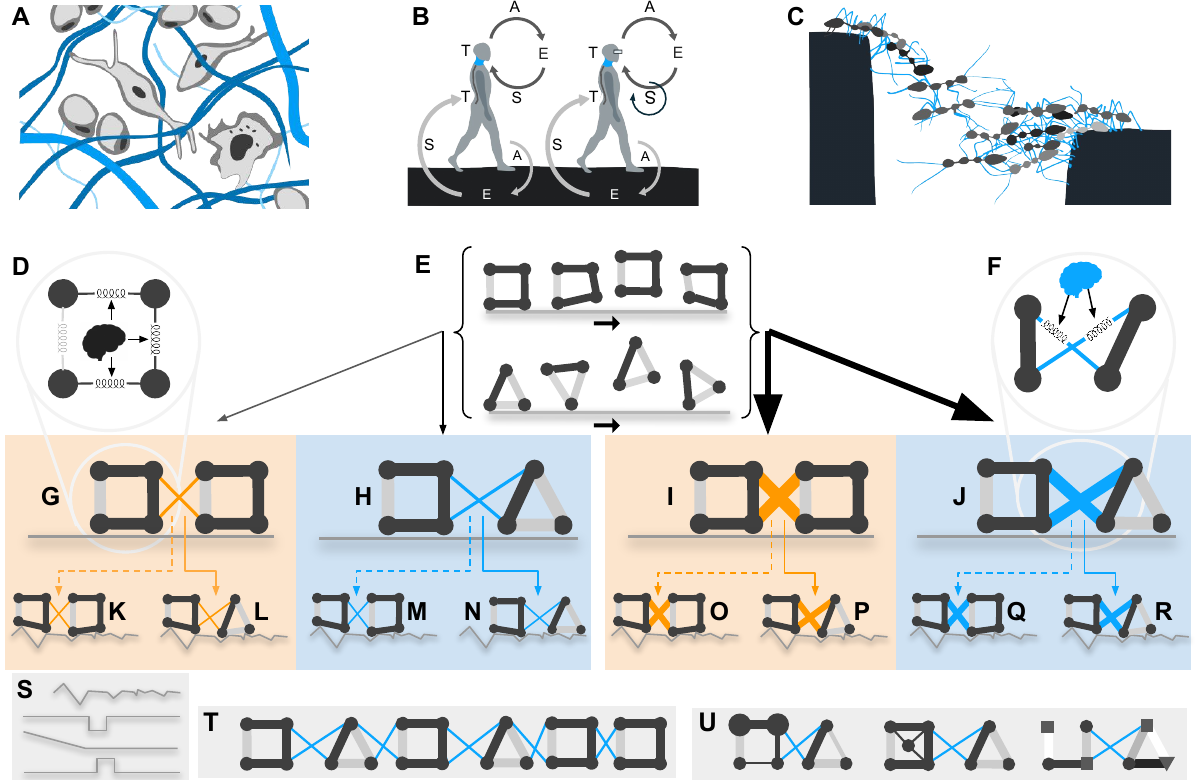}}
\caption{
%
\textbf{Internal diversity across scales in nature.}
A cell’s (\textbf{a}), anatomical joint’s (\textbf{b}), and organism’s (\textbf{c}; blue) experiences are broadened by diverse cell types, changing relative body part sizes, and different caste members encountered during cell migration, development, and locomotion, respectively. This may help damaged cells, visual inversion via prismatic glasses, or a dangling conspecific (red) fall within the cell's, joint's, or organism's zones of competence, allowing them to extinguish rather than spread surprise. 
\textbf{Internal diversity in machines.} 
(\textbf{d}) diverse agents built from active (black) and passive (gray) springs were trained for forward motion (\textbf{e}). Connectors also made of springs (\textbf{f}) were then tethered to and untethered from a series of agent pairs, and rewarded for causing the resulting collectives to move forward. Weak connectors were so trained on agents and their clones (\textbf{g}) or different agents (\textbf{h}), as were strong connectors (\textbf{i,j}), resulting in four sets of connectors.  Finally, collectives containing connectors trained and/or deployed within uniform or diverse collectives (\textbf{k-r}) were deployed to new environments. We found that weak connectors trained and deployed within diverse collectives (n) yielded the most resilient collectives, across many novel environments (\textbf{s}). Loosely coupling more (\textbf{t}) or more diverse (\textbf{u}) agents resulted in even greater resilience.
}
\label{fig:1}
\end{figure*}

These many different architectures share a common function: they allow the whole system to adapt rapidly to new challenges by containing local failures or missteps without suppressing individual variability \cite{glassman1973persistence,kitano2004biological,almaatouq2020adaptive}. They balance independence and cooperation through modular connections and selective integration, so that diversity becomes a source of resilience rather than fragility \cite{kitano2004biological}. Some representative biological examples are visualized in Fig. \ref{fig:1}.

The value of diversity has long been recognized in AI and robotics. In machine learning, ensemble methods that combine diverse models improve robustness and out-of-distribution generalization \cite{58871,dietterich2000ensemble}. In robotics, projects like Swarmanoid have used heterogeneous groups of specialized robots for complex tasks like search-and-retrieval~\cite{dorigo2013swarmanoid}, while experiments with swarms of non-calibrated robots have shown that persistent individual differences—often dismissed as noise—can be potential assets \cite{raoufi2023individuality}. 

AI and robotics also frequently make use of loose coupling: modularity is a hallmark of both hardware and software design~\cite{mammela2023loose}. In AI, Mixture-of-experts architectures selectively route information through sparsely connected expert modules~\cite{masoudnia2014mixture,DBLP:conf/iclr/ShazeerMMDLHD17}, and research in swarm and modular robotics has explored how communication constraints and decentralized control can serve as technological analogues of certain types of natural collective coupling~\cite{baca2012heterogeneous,talamali2021less,aust2022hidden,zakir2024heterogeneity,hepworth2024contextually}.

Even so, most engineered collectives still emphasize homogeneity \cite{dorigo2013swarmanoid,9460560,10.1162/ARTL_a_00116}, centralized or tightly coordinated control~\cite{schranz2020swarm}, and operation within well-defined task boundaries~\cite{DBLP:journals/corr/abs-2501-13075}. The interplay between heterogeneity and loose coupling and its role in supporting robustness in novel situations remains relatively unexplored. In this study, we set out to examine, in a minimal computational framework, how resilience can emerge specifically from combining morphological diversity and trainable, distributed coupling mechanisms, without the benefit of pre-adaptation to a range of environments or reliance on hand-designed division of labor (see Fig. \ref{fig:1}).

\section{Approach.}

To explore how internal diversity can prepare organisms or machines for external surprise, we first train diverse components to exhibit an arbitrarily-chosen desired behavior: rightward motion. We then tether different components together with connectors, thereby creating internally diverse collectives. The connectors learn to enable the paired components to move together, resulting in collectives that also exhibit rightward motion. Finally, we deploy the collectives to novel environments to see which can still move there.

\subsection{Components.}

We begin by automatically generating a population of simulated components, each of which exhibits the desired behavior of locomotion on its own, in a unique way.

We accomplish this by first creating 100 random components in simulation. Each is morphologically unique: it is composed of 11 point masses placed at random relative locations and connected together by 21 randomly-chosen passive or active springs (Fig. \ref{fig:1}D). Each component is also neurologically unique: at each simulation time step, external sensation and internal sine waves are supplied to the component's neural controller, which mixes them together in a unique way. The controller then dictates length changes to the active springs. 

Each component is simulated 15 times on flat ground. After each simulation, the gradient descent optimization method \cite{degrave2019differentiable, hu2020diffTaichi} slightly alters how the neural controller mixes that component's sensation and internal sine waves to produce action, and alters the positions of its point masses, to ensure that during the next simulation that component exhibits slightly further displacement.

An evolutionary algorithm \cite{sims1994evolving,lipson2000automatic,floreano2008evolutionary} then takes over from the gradient descent algorithm. The evolutionary algorithm uses the body plans and final displacements of all 100 components to delete the less morphologically unique and slower components, and to make randomly-modified copies of the surviving more morphologically unique and faster components, following \cite{lehman2011evolving}. During each copying event, some randomly-chosen springs are toggled from active to passive, or vice versa. Also, some of the positions of its point masses are randomly changed. 

Gradient descent then takes over again to tune the bodies and brains of each new offspring, after which the evolutionary algorithm again discards less-unique and slower components and spawns offspring from the more unique, faster components. This alternation follows \cite{matthews2023efficient} who showed that this approach allows the evolutionary algorithm to efficiently explore body plans, while gradient descent can carefully tune each brain to best exploit that body's behavioral potential, producing at the end 100 competent and diverse components (Fig. S1). While each is competent in locomotion, their different body designs naturally lead them to adopt different movement strategies (Fig. \ref{fig:1}E). This behavioral diversity becomes an important resource later, when components are brought together into collectives.

\subsection{Connectors.}
We next trained connectors (Fig. \ref{fig:1}F) that tether pairs of these independently capable components together in different ways. Every connector has five actuated springs, with endpoints algorithmically chosen from key points on each component’s body based on that component's movement and ground contact patterns. This ensures the connector attaches at locations likely to facilitate coordinated movement, such as connecting the estimated front of the posterior component to the estimated back of the anterior component, regardless of component morphology. Some components roll partially, complicating the identification of its front or back.

We trained four types of connectors:
10 that loosely bind a component to its clone (Fig. \ref{fig:1}G),
10 that loosely bind two different components (Fig. \ref{fig:1}H),
10 that tightly bind a component to its clone (Fig. \ref{fig:1}I), and
10 that tightly bind two different components (Fig. \ref{fig:1}J).
Each of these 40 connectors begins training with a random neural controller that randomly transforms sensed features of its two components (such as their relative positions)
into actions of its five springs (lengthening or shortening).

The first connector from the first type (Fig. \ref{fig:1}G) tethered the first component to its clone, and the resulting uniform collective was simulated.
We disconnect it and then tether it to the second component and its clone, and the new collective was simulated.
This process was repeated for the remaining 98 uniform collectives.
The degree to which the connector disrupted the 200 components' rightward travel was used to tune the connector's neural controller using gradient descent such that when it tethered the 100 component pairs together again, the 100 collectives traveled further to the right on average.
This training process was repeated 100 times.
The components' controllers were not modified during this process.
The remaining nine connectors in the first set were trained in the same manner. 
The 10 strong connectors within uniform collectives (Fig. \ref{fig:1}I) were also trained this way.

The first connector in the second set (Fig. \ref{fig:1}H) was trained by 
using it to tether two randomly-chosen components together, simulating the resulting diverse collective,
and repeating this for 99 more pairs of randomly-chosen components.
Again, the degree to which the connector disrupted the 200 components' rightward travel was used to tune the connector's neural controller using gradient descent.
This training process was repeated 99 more times, 
and this connector training regimen was in turn repeated nine more times for the remaining nine connectors of this type.
Finally, the 10 strong connectors within diverse collectives (Fig. \ref{fig:1}J) were also trained this way.

\subsection{Collectives.}
To assess whether internal diversity and loose coupling confers the ability to grapple with external surprise,
eight types of collectives were deployed into new environments.
The first four collective types contained weak connectors that 
never experienced component diversity                          (Fig. \ref{fig:1}K),
experienced component diversity only during deployment         (Fig. \ref{fig:1}L),
experienced component diversity only during training           (Fig. \ref{fig:1}M), or
experienced component diversity during training and deployment (Fig. \ref{fig:1}N).
The fifth through eighth collective types contained strong connectors that
never experienced component diversity                          (Fig. \ref{fig:1}O),
experienced component diversity only during deployment         (Fig. \ref{fig:1}P),
experienced component diversity only during training           (Fig. \ref{fig:1}Q), or
experienced component diversity during training and deployment (Fig. \ref{fig:1}R).
During deployment to new environments, no further component or connector training was allowed: 
we simply observed how well or poorly the collectives grappled with the resulting surprise.

\subsection{Environments.}
All collective types were then deployed into 12 novel environments (Fig. \ref{fig:1}S). Each environment differs from the flat terrain used during training and is designed to pose a unique challenge to locomotion. Some involve changes in ground geometry, such as slopes, steps, gaps, or obstacles. Others alter surface properties or introduce external forces, including adhesive ground, ice, sand, wind, treadmill motion, and reduced gravity. While not exhaustive, these scenarios are intended to represent a broad range of conditions that natural or artificial collectives might encounter.

A final, two collective types were deployed into these environments: 
collectives with more components   (Fig. \ref{fig:1}T) and
those with more diverse components (Fig. \ref{fig:1}U).

\begin{figure*}[ht!]
\centering
\includegraphics[width=.98\linewidth]{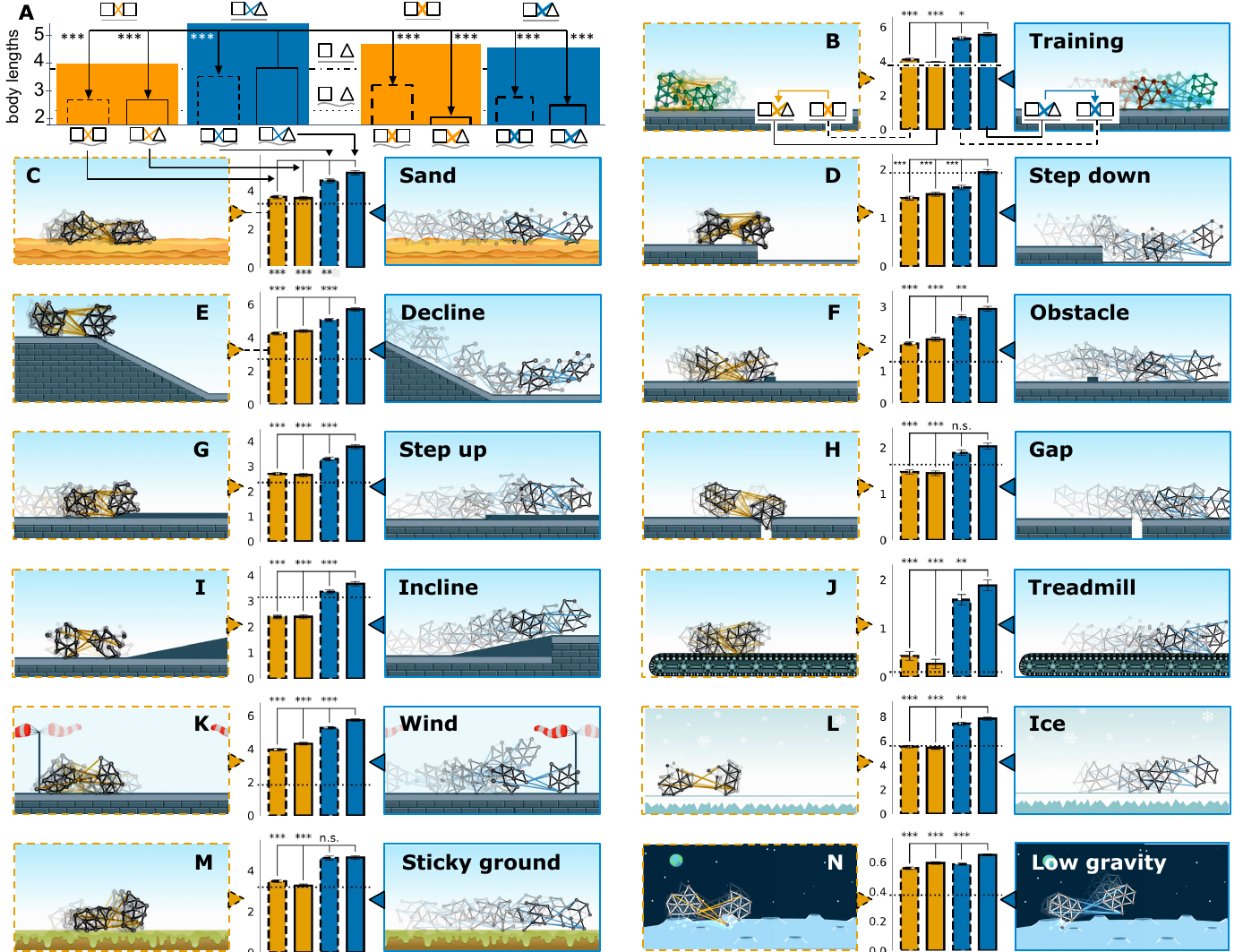}
\caption{
\textbf{Internal diversity and loose coupling confer resilience.} 
\textbf{a}, Mean forward motion of independent agents trained on flat ground (dashed dotted)
and when deployed to 12 new environments (dotted).
Mean forward travel achieved by weak connectors (left two large bars) trained within uniform (orange) or diverse (blue) collectives and strong connectors (right two large bars) trained within uniform (orange) or diverse (blue) collectives, on flat ground.
Mean forward travel of these four sets of connectors when deployed to the 12 environments within uniform collectives (inset dashed bars) or diverse collectives (inset solid bars).
Error bars are visually unresolvable.
\textbf{b}, forward travel caused by weak connectors when 
trained in uniform collectives                                (orange dashed) and then 
`deployed' within diverse collectives back on flat ground (orange solid),
and forward travel caused by weak connectors when 
trained in diverse collectives                                (blue solid) and  then 
`deployed' within uniform collectives back on flat ground (blue dashed).
\textbf{c-n}, Mean forward travel caused by the four sets of weak connectors that
did or did not experience diversity during training and deployment, for each environment.
Representative behaviors are shown for a connector that never experienced diversity (left image) or experienced it during training and deployment (right image).
}
\label{fig:2}
\end{figure*}

\section{Results.}

Overall, it was found that independent components suffered the greatest average decrease in performance when deployed to new environments (Fig. \ref{fig:2}A dashed-dotted and dotted lines), 
while collectives containing 
weak connectors that experienced diversity during training and deployment enabled their collectives to suffer the least (Fig. \ref{fig:2}A thin solid blue bar).

\subsection{Component training.}

Independent components were successfully trained on flat ground: trained components moved on average almost four body lengths to the right, while random ones averaged zero body lengths (Fig. \ref{fig:2}A, dashed-dotted line).
When deployed, they were severely compromised, moving just more than two body lengths to the right on average across all 12 novel environments (Fig. \ref{fig:2}A, dotted line).

\subsection{Connector training.}

Tethering trained components with initially untrained connectors led to severe drops in behavior (not shown),
but subsequent connector training yielded collectives that traveled further than the independent components, 
for all four types of collectives (Fig. \ref{fig:2}A, four large bars). 
Collectives with internal diversity and loose coupling (Fig. \ref{fig:2}A, left large blue bar) 
achieved the best performance, suggesting that connectors within those collectives learned to
(1) rectify the behavior of those components misbehaving due to their lost independence;
(2) not disrupt the behavior of those components that suffered little from being tethered; and
(3) exploit the collective's internal diversity for faster travel.

\subsection{Collective deployment.}

When the trained components and trained connectors were recombined into the eight collective types and deployed to the novel environments, performance for all eight collective types drops, averaged across all 12 environments (Fig. \ref{fig:2}A, eight small bars).
However, weak connectors which experienced diversity during training and deployment enabled their collectives to fare best
(Fig. \ref{fig:2}A, small blue thin-outlined bar). This suggests that these connectors' ability to exploit internal diversity to confer performance on the collective in the training environment was not lost when faced with external novelty.

Fig. \ref{fig:2}B sheds more light on the advantage enjoyed by weak connectors trained within diverse collectives (solid blue bar).
When they are transplanted into uniform collectives,
those collectives show only a small decrease in performance (dashed blue bar), 
indicating they have also become adept at handling component clones, a situation they never encountered during training.
Conversely, weak connectors trained within uniform collectives (dashed orange bar) are not able to exploit 
diversity when they are transplanted into diverse collectives (solid orange bar).

Given the poor performance of collectives deployed with strong connectors (Fig. \ref{fig:2}A rightmost four small bars), 
Fig. \ref{fig:2}C-N reports the relative performance of only the four
loosely-coupled collective types (also, in all subsequent analyses, we focus on the condition using loose connectors).
Again, in most environments, collectives with weak connectors that experienced diversity during 
training and deployment (solid blue bars) fared best.
They also fared much better than the independent components operating in those environments (dotted lines).
They also, in some cases, fared better than collectives trained just in one of these environments (Fig. S2).

\section{Discussion.}

\subsection{Exploiting internal diversity.}

\begin{figure*}[!t]
\centering
\includegraphics[width=.95\linewidth]{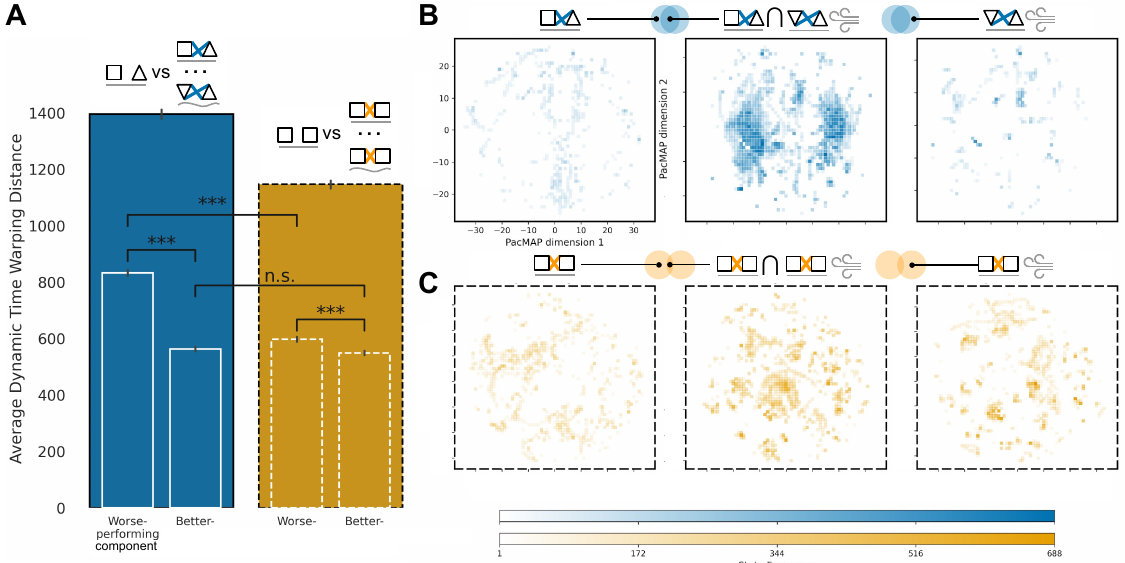}
\caption{%
%
\textbf{Internal adversity.} 
\textbf{a}, Connectors within diverse collectives experienced greater internal adversity than those within uniform collectives: Tethered agents within diverse collectives (blue) were disrupted more than agent clones within uniform collectives (orange). (Mean dynamic time warping distance (DTWD) was used to quantify how much a tethered agent’s behavior differed from its independent behavior, after the connector training.) In general, adversity was tamed: worse-behaving agents were altered more than well-behaving agents in both collectives. But, diversely-trained connectors tamed adversity more: diversely-trained connectors disrupted worse-behaving agents more than uniformity-trained connectors did (left white bar and white-dashed bar) but had the same mild impact on the better-behaving agents. (Error bars denote bootstrapped 95\% confidence intervals around 10 connectors, each trained on 100 agent pairs. $^{***} = p < 0.001$).
\textbf{External adversity.}
\textbf{b}, A representative diversity-trained connector received many more percepts that arose in both the training and wind environments (middle panel) compared to percepts that only occurred to it in one of those environments. (Each point represents a percept and opacity encodes how many times that percept occurred.) 
\textbf{c}, In contrast, a representative uniformity-trained connector had many more percepts unique to the training or wind environment, indicating wind felt
more unfamiliar to it than wind did to the diversity-trained connector.}
\label{fig:3}
\end{figure*}

The question arises as to why internally diverse collectives do so well.
It could be that diverse components simply act as a diverse portfolio of behaviors, increasing the likelihood that at least one component in the collective will be unsurprised by external novelty. If this were true, connectors should learn to interfere as little as possible with component operation. Fig. \ref{fig:3}A however suggests that this is not what occurs. 

Instead, we found that weak connectors learn to affect the worse- and better-performing components within a collective differently. 
(The better-performing component is defined as the one that moved further right on its own, compared to the other component on its own.)
Weak connectors trained within diverse collectives (large blue bar) and 
those trained within uniform collectives (large orange bar)
altered the behavior of the worse-performing component more than they altered the behavior of the 
better-performing component,
on average,
for both collective types,
across the training environment and 12 novel environments
(compare the two small blue bars and compare the two small orange bars).
Moreover, diversely trained connectors caused a significantly larger change in the worse-performing component's behavior than uniformly trained connectors did 
(compare left small blue bar and left small orange bar),
while the connectors' effect on better-performing components was comparable
(compare right small blue bar and right small orange bar).
This shows that diversely-trained connectors do not learn to avoid disrupting component behavior. 
Instead, 
they learn to adapt better to the behavioral differences that naturally arise in novel conditions and can more effectively mediate between components with distinct movement patterns.

To further probe why internally-diverse collectives fare better when confronting novelty,
we analyzed the perceptual states connectors attain in both training and unfamiliar environments. We did so using PaCMAP, a nonlinear dimensionality-reduction technique designed to preserve both local and global structure in high-dimensional data \cite{wang2021understanding}. By mapping the inputs arriving at one diversely-trained and one uniformity-trained connector into two dimensions, we were able to approximate and visualize the states those two connectors encountered during training and when encountering a novel environment.
For the diversely-trained connector, there was substantial overlap: many of the states it encountered in the new environments were similar to those it had seen during training, essentially making the novel environment appear more familiar
(Fig. \ref{fig:3}B). 
In contrast, uniformly trained connectors encountered many states that were unique to the new environment and had little overlap with training
(Fig. \ref{fig:3}C). 
This pattern held across most environments, component pairs, and connectors (Table S1). 
This indicates that new environments `feel' more familiar to internally-diverse collectives than they do to 
internally-uniform collectives.

Taken together, these findings suggest a two-fold benefit from diversity. First, training connectors on diverse component pairs enables them to learn mediating strategies that generalize better to novel patterns of behavior and performance. Second, assembling collectives from diverse components ensures a broader repertoire of responses to unexpected challenges, which connectors can then exploit to sustain group function. 
This results in the fact that, compared to connectors trained without diversity,
the broader set of experiences encountered and tamed by connectors trained with diversity
overlaps more with the experiences they encounter when their collectives encounter novelty, allowing diversity-trained connectors to tame those seemingly novel situations better.

\subsection{Scaling diversity.}

\begin{figure}[!t]
\centering
\includegraphics[width=1\textwidth]{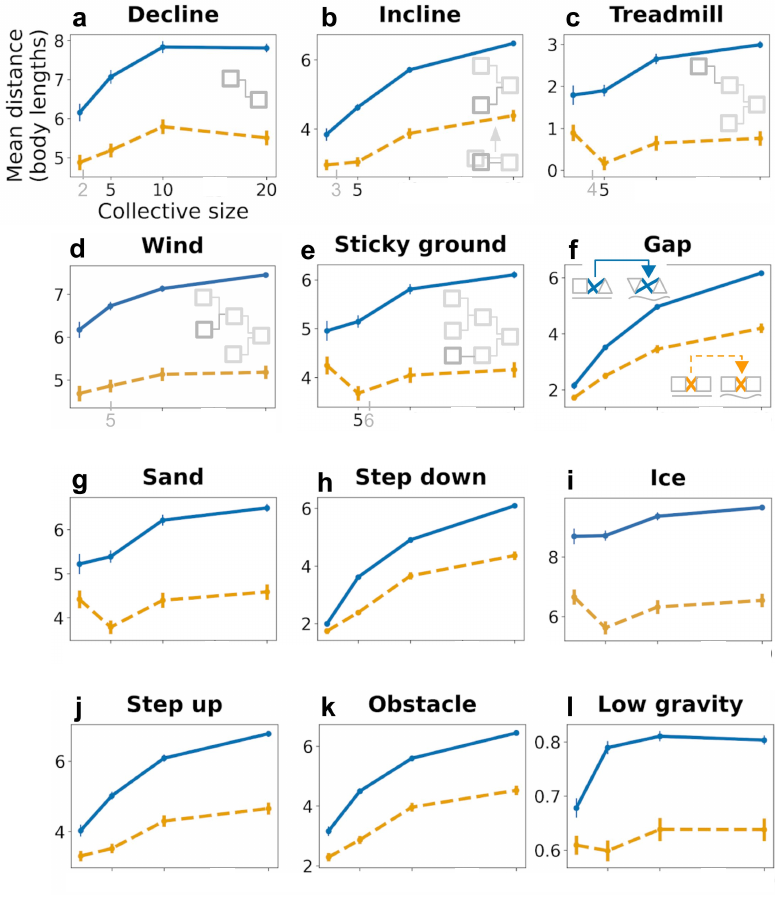}
\caption{\textbf{Performance of uniform and diverse collectives across environments and collective sizes.}  
Orange dashed lines show results for weak connectors trained within uniform two-agent collectives 
but deployed within larger, uniform collectives. 
Blue lines          show results for weak connectors trained within diverse, two-agent collectives 
but deployed within larger, diverse collectives. 
Vertical error bars represent 95\% confidence intervals.
}
\label{fig:4}
\end{figure}

To investigate whether internal diversity also confers resilience as the collective scales up in size and complexity, 
we connecting together copies of already-trained components and connectors to create five, 10, and 20-component collectives. Components were chained together in a branching structure: a given component could have zero, one, or two other components connected to the left of it. Examples of this component chaining for two- through six-component collectives are shown as gray insets in Fig. \ref{fig:4}. 
We created 1000 unique five, 10 and 20-component uniform collectives containing uniformity-trained connectors (Fig. \ref{fig:4}, orange dashed). We also created 1000 unique five, 10 and 20-component diverse collectives containing diversity-trained connectors (Fig. \ref{fig:4}, solid blue). 
Each of these collectives was deployed into each of the novel environments.
As before, their ability to move right in these new situations, without the benefit of further training, was measured.
See the Scaling Diversity section in \textit{SI supporting text} for more details of this construction process.

We found that the performance advantage of internal diversity grows with collective size. Further, this advantage comes without any additional training: it simply requires chaining more diverse components and more diversity-trained connectors together. Finally, it demonstrates another ability of diversity-trained connectors: even when trained within two-component collectives in one environment, they adapt better than uniformity-trained connectors when they find themselves within larger collectives, in new environments. The Morphological Diversity section in \textit{SI supporting text} reports another way in which increased internal diversity confers increased resiliency: as components differ in more ways, the collectives they reside within exhibit more resiliency.

\subsection{Biological implications.}

Our results support the increasingly recognized view in biology that within-group diversity is not merely noise but serves a  functional role~\cite{o2021functional}. 
Specifically, we have shown how this can occur among mechanically coupled group members: 
diverse components provide broad experiences to their connectors, 
so much so that encounters with new environments trigger local behaviors which fall within the connectors' ability to recover operation therein.
We do not claim that the mechanisms described in our models map directly onto particular biological cases, nor that diversity and coupling are always beneficial or operate in the same way wherever they appear. 
Instead, our work suggests that organisms, under some conditions, may be actively
conducting internal mechanical drills, using internal diversity, 
to prepare for external adversity.
Further, this pre-gaming may be occurring across subcellular through ecological scales.

\subsection{Technological implications.}

With only a few exceptions~\cite{li2019particle}, 
current robots and autonomous vehicles lack interiors with semi-independent, 
functionally overlapping parts, disbarring them from this opportunity to prepare within for surprise without.
Our approach is also orthogonal to the currently prevailing closed-world assumption of machine learning
and robotics,
which is that the best way to ensure reliable operation in new environments is
to train it in enough environments that some will
resemble those it will encounter during deployment.
Of course, increasing the diversity of a collective brings with it practical engineering challenges. 
It can require substantial design effort to create connective architectures that can support coordination and communication across dissimilar components with a wide range of morphologies and capabilities.
But, scalable automated robot design~\cite{matthews2023efficient} offers the prospect of creating large numbers of 
mechanically unique robots,
and advances in modular robotics~\cite{savoie2019robot} are facilitating the housing of smaller robots within larger ones.

\section{Conclusions.}

While our results show the value of internal diversity and mechanical coordination, they are based on idealized simulations with limited types of component variation and coupling. Real-world biological and technological systems face many additional sources of complexity—energy constraints, hierarchical structure, active communication, and ongoing adaptation that remain outside our present scope. Exploring these factors, and developing machines that enjoy richer inner dynamics, remain important directions for future research.
Nonetheless, our results suggest that organisms and machines that not only contain but actively exploit their rich inner lives
may help explain biological robustness
and eventually yield machines capable of reliable operation in an open, unpredictable world.

\section{Funding Statement.}

This work was supported by Army Research Office contract W911NF-23-1-0327.

\bibliography{DivColl}
\bibliographystyle{naturemag}

\newpage

\section{Supplementary Information}

\subsection{Supplementary Methods}

\subsubsection{Agents.}
Each agent consists of 11 point masses ($n_{PM}$) connected by 21 springs in a triangular grid of $4$ rows $\times$ $7$ columns. 
Its morphology is defined by a ``body genotype'' (a binary vector specifying occupied locations) and ``spring genotype'' (a binary vector denoting which springs are active). Upon generation, body genotypes are post-processed with a flood-fill procedure to remove disconnected agents and fill internal holes to ensure structural integrity. All springs have a fixed rest length of $0.05$ length units (the distance between two adjacent points in the grid) and the same stiffness; active springs all have the same actuation amplitude.
Each agent is controlled by a two-layer feedforward neural network. The network receives periodic timing signals (10 sine waves) and proprioceptive features: the velocities and relative positions (with respect to the component’s center of mass) of all mass points. Thus, the input dimensionality is $10+4n_{PM}=54$. The network’s two layers each have $32$ hidden units and use a $\tanh$ activation function. Network outputs determine the change to the rest length of each spring. To ensure stable actuation, each network output at timestep $t$ ($o_t$) is computed as a weighted combination of its current and previous value: $o_{t} = 0.15\,o_{t} + 0.85\,o_{t-1}$.
Agent performance is quantified by the net forward displacement of its center of mass over the simulation:
\begin{equation}
\text{fitness} = x_{\mathrm{COM}}(T) - x_{\mathrm{COM}}(0)
\end{equation}
where $x_{\mathrm{COM}}(t)$ is the center of mass horizontal coordinate at time~$t$ and $T$ is the final timestep.

\subsubsection{Agent optimization.} 
We employed an evolutionary algorithm to optimize agent bodies and gradient descent to optimize the neural network controller of each body.
The evolutionary algorithm begins with 100 agents with random bodies and neural network controllers.
The neural networks of each of these agents, and the positions of their point masses,
is optimized via 15 epochs of gradient descent, in simulation,
against the performance metric above.
Gradients are clipped at $0.16$. 
We use an adaptive learning rate that is scaled according to the root-sum-of-squares of all weight gradients. 
To ensure agent diversity, we compute each component's 
normalized height and width, aspect ratio,
bounding box compactness (active springs per area), centroid
position, standard deviation of vertical mass point locations
(balance), and measures of vertical and horizontal reflectional
symmetry. 
The resulting vector describes a component's position in morphospace.
We assign each agent a uniqueness score, computed as its average
distance from its five nearest neighbors in this space.
Each agent then produces one mutated offspring.
During mutation, bits in both body and spring genotypes are flipped with a probability of $1/L$ (where $L$ is genotype length). 
Constraints are enforced to ensure that the target number of point masses is met. 
The population of 200 agents is now reduced by deleting the agent with the lowest
fitness times uniqueness score until 100 remain, concluding the generation.
Fourteen more generations are performed, which yields 100 fit and morphologically diverse components.

\noindent \textbf{1.2.1 The forward pass.} 
Each of the 15 forward passes for an agent involved simulating it for 1000 time steps.
During those time steps, gradients were computed at various points within the body and neural control policy of the component.

\noindent \textbf{1.2.2 The backward pass.} 
When the simulation of an agent terminates, 
errors flow backward 
from the agent's final horizontal displacement to the final positions of its nodes, then
on to the final velocities of its nodes, then
onward to the forces applied to those nodes by their attached springs,
then onward to the accelerations of those springs,
then onward to the desired lengths sent to those springs by the output neurons of the component's neural control policy,
then onward to the weights of the policy,
then onward to the input values to that policy,
then onward to the sensor values of the agent that populate its sensors,
then the positions of nodes that determine those sensor values.

\noindent \textbf{1.2.2.1 Agent position to node positions.}
Let $L$ be the loss, and $x^{\mathrm{agent}}_T$ the final center-of-mass $x$ position after $T$ steps. The loss gradient is:
\begin{equation}
\frac{\partial L}{\partial x^{\mathrm{agent}}_T} = 1
\end{equation}
$N$ is the number of nodes in the agent. If the agent's $x$ coordinate is computed as the mean of its node positions:
\begin{equation}
x^{\mathrm{agent}}_T = \frac{1}{N} \sum_{i=1}^N x^{(i)}_T
\end{equation}
Hence, for each node $i$,
\begin{equation}
\frac{\partial L}{\partial x^{(i)}_T} = \frac{1}{N}
\end{equation}
The loss signals are then propagated to each node's position at the final time step.

\noindent \textbf{1.2.2.2 Node positions to node velocities.}
Let $\mathbf{x}_t^{(i)}$ and $\mathbf{v}_t^{(i)}$ be the position and velocity of node $i$ at time $t$.
Node positions are updated by:
\begin{equation}
\mathbf{x}_{t+1}^{(i)} = \mathbf{x}_t^{(i)} + \mathbf{v}_t^{(i)} \Delta t
\end{equation}
where $\Delta t$ is the simulation timestep.
During the backward pass, the gradient with respect to $\mathbf{v}_t^{(i)}$ accumulates from its contribution to $\mathbf{x}_{t+1}^{(i)}$ via:
\begin{equation}
\frac{\partial L}{\partial \mathbf{v}_t^{(i)}} \gets \frac{\partial L}{\partial \mathbf{v}_t^{(i)}} + \Delta t \cdot \frac{\partial L}{\partial \mathbf{x}_{t+1}^{(i)}}
\end{equation}
Additionally, the loss gradient is passed back to $\mathbf{x}_t^{(i)}$:
\begin{equation}
\frac{\partial L}{\partial \mathbf{x}_t^{(i)}} \gets \frac{\partial L}{\partial \mathbf{x}_t^{(i)}} + \frac{\partial L}{\partial \mathbf{x}_{t+1}^{(i)}}
\end{equation}

\noindent \textbf{1.2.2.3 Node velocities to node forces.}
Velocities of each node are updated according to Newton's second law:
\begin{equation}
\mathbf{v}_{t+1}^{(i)} = \mathbf{v}_t^{(i)} + \frac{\mathbf{F}_t^{(i)}}{m_i} \Delta t
\end{equation}
where $\mathbf{F}_t^{(i)}$ is the total force on node $i$, and $m_i$ its mass.
The gradient with respect to $\mathbf{F}_t^{(i)}$ at time $t$ is:
\begin{equation}
\frac{\partial L}{\partial \mathbf{F}_t^{(i)}} \gets \frac{\partial L}{\partial \mathbf{F}_t^{(i)}} + \frac{\Delta t}{m_i} \frac{\partial L}{\partial \mathbf{v}_{t+1}^{(i)}}
\end{equation}
The gradient with respect to velocity at the previous step is:
\begin{equation}
\frac{\partial L}{\partial \mathbf{v}_t^{(i)}} \gets \frac{\partial L}{\partial \mathbf{v}_t^{(i)}} + \frac{\partial L}{\partial \mathbf{v}_{t+1}^{(i)}}
\end{equation}

\noindent \textbf{1.2.2.4 Node forces to spring forces.}
The total force $\mathbf{F}_t^{(i)}$ on each node is decomposed into spring forces, actuation, damping, gravity and contact forces. For each spring connecting nodes $i$ and $j$, the spring force (with actuation) is:
\begin{equation}
\mathbf{f}_{\text{spring}}^{(i,j)} = k_{ij} \left( \| \mathbf{x}_t^{(i)} - \mathbf{x}_t^{(j)} \| - l_{0,ij}(1 + a_{ij} \, u_{ij,t}) \right) \hat{d}_{ij} - d_{ij} (\mathbf{v}_t^{(i)} - \mathbf{v}_t^{(j)})
\end{equation}
where:
\begin{itemize}
    \item $k_{ij}$ is the spring stiffness,
    \item $l_{0,ij}$ is the nominal (unactuated) spring rest length,
    \item $a_{ij}$ is the actuation scaling parameter for the spring,
    \item $u_{ij,t}$ is the neural network output (actuation command) for the spring at time $t$,
    \item $\hat{d}_{ij} = \frac{\mathbf{x}_t^{(i)} - \mathbf{x}_t^{(j)}}{\|\mathbf{x}_t^{(i)} - \mathbf{x}_t^{(j)}\|}$ is the unit direction vector from node $j$ to node $i$,
    \item $d_{ij}$ is the spring damping coefficient.
\end{itemize}
The gradient $\frac{\partial L}{\partial \mathbf{f}_{\text{spring}}^{(i,j)}}$ is backpropagated analytically through the force equation to relevant node positions, velocities, and actuation inputs.

\subsubsection{Connectors.}

Each connector comprises five springs, joining specific pairs of agent points as determined by pre-identified anchor indices. Anchor points are automatically assigned: we select two ``feet'' (points with the most frequent ground contact on either side) and two stable upper points (the highest lateral mass points exhibiting the least positional variability across a simulation on flat ground). Connector springs are assigned a rest length equal to the anchor point pair distance at initialization, and an actuation amplitude of $0.6$ unit length. 
Weak and strong connectors contain springs with low and high stiffness respectively.

Connectors are also governed by a neural network controller. This controller is cloned within all connectors, 
in all collectives of the same type.
This controller receives $4+4+2+1=11$ input features per spring:
endpoint positions relative to connector's midpoint, 
endpoint velocities, 
touch sensor signals at both endpoints, and 
current spring length. 
The neural network consists of two layers~(each with $32$ hidden units, $\tanh$ activation), and outputs change the resting length of each connector spring at every timestep. Connector outputs are also smoothed over time: $o_{t} = 0.1o_{t} + 0.9o_{t-1}$. For the visualization in Fig.~\ref{fig:3}B,C, the $32$-dimensional hidden state activations of connector controllers were recorded during simulation across all $100$ agent pairs and projected into two dimensions using PaCMAP. PaCMAP embeddings were computed using default parameters: $10$ nearest neighbors, a mid-near pair/neighbor ratio of $0.5$, and a further pair/neighbor ratio of $2$.

\subsubsection{Connector optimization.}
Connector neural network weights were trained using backpropagation over batches of collectives, where each collective comprised two agents and one connector (each connector composed of a set of springs). Unlike agents, connector morphology was hand-designed, and a single set of policy weights was shared across all collectives in the batch.

\noindent \textbf{1.4.1 The forward pass.}
For each batch, every collective was simulated for $T$ time steps. At each step, the connector's neural network received sensor inputs and outputted actuation commands $u_{ij,t}$ for each connector spring $(i,j)$. These actuation commands modulated spring lengths, influencing the positions of connected nodes.

\noindent \textbf{1.4.2 The backward pass.}
At the end of the simulation, gradients flowed backward from the collectives' displacement through the connector springs to the connector neural network weights. The process unfolded as follows.

\noindent \textbf{1.4.2.1 Collective displacement to node positions.} 
Similar to agent optimization, the collective loss $L$ is based on the center-of-mass $x$ displacement across all point masses in both agents:
\begin{equation}
\frac{\partial L}{\partial x^{\mathrm{collective}}_T} = 1
\end{equation}
Assuming $N$ masses total, with $x^{\mathrm{collective}}_T = \frac{1}{N} \sum_{i=1}^N x^{(i)}_T$, the gradient for each mass $i$ is:
\begin{equation}
\frac{\partial L}{\partial x^{(i)}_T} = \frac{1}{N}
\end{equation}

\noindent \textbf{1.4.2.2 Mass point positions to connector spring forces.} 
Mass point positions are determined in part by the connector spring forces. For each connector spring $(i,j)$,
\begin{equation}
\mathbf{f}_{\text{conn}}^{(i,j)} = k_{ij} \Big( \| \mathbf{x}_t^{(i)} - \mathbf{x}_t^{(j)} \| - l_{0,ij}(1 + a_{ij} \, u_{ij,t}) \Big) \hat{d}_{ij} - d_{ij} (\mathbf{v}_t^{(i)} - \mathbf{v}_t^{(j)})
\end{equation}
The loss gradient is analytically backpropagated through this force equation to the relevant variables (point mass positions and actuation commands).

\noindent \textbf{1.4.2.3 Connector spring forces to actuation commands.} 
The neural output $u_{ij,t}$ determines the actuation of spring $(i,j)$ at time $t$. The gradient passed to each $u_{ij,t}$ is:
\begin{equation}
\frac{\partial L}{\partial u_{ij,t}} = - a_{ij} \, k_{ij} \, l_{0,ij} \, \hat{d}_{ij} \cdot \frac{\partial L}{\partial \mathbf{f}_{\text{conn}}^{(i,j)}}
\end{equation}

\noindent \textbf{1.4.2.4 Actuation commands to connector network weights.} 
Each actuation $u_{ij,t}$ is the output of the connector neural network given its inputs at that step:
\begin{equation}
u_{ij,t} = \mathrm{NN}_{\text{conn}}(\mathbf{s}_{ij,t}; \, \mathbf{w})
\end{equation}
The neural network's parameters $\mathbf{w}$ are shared across the batch, and gradients are accumulated:
\begin{equation}
\frac{\partial L}{\partial \mathbf{w}} = \sum_{b=1}^B \sum_{t=1}^T \sum_{(i,j) \in \text{conn}_b} \frac{\partial L}{\partial u_{ij,t}} \cdot \frac{\partial u_{ij,t}}{\partial \mathbf{w}}
\end{equation}
where $B$ is the batch size and $\text{conn}_b$ is the set of connector springs in collective $b$. Importantly, the agents' neural network controllers are not modified. Only the connector's policy is updated.

\subsubsection{Simulation.}

Simulations were conducted using a custom physics engine built with Taichi ($\Delta t = 0.004s$). Training episodes lasted $1{,}000$ steps, and deployment episodes ran for $3{,}000$ steps. Each environment applied default downward gravity ($-4.8\frac{m}{{s}^{2}}$), friction ($1.0$), and damping ($15$), unless noted otherwise.

Terrain was modeled as arrays of ground segments, each with configurable slopes and elevations. The training terrain consisted of a uniform, horizontal segment. Friction-modified environments included ``ice'' (friction $1.5$) and ``sticky'' (friction $0.85$). ``Wind'' environments introduced a sinusoidal lateral force, amplitude $0.04\frac{m}{s}$, alternately changing direction during an episode. ``Sand'' environments used a damping value of $20$, increased energy loss on ground contact (absorption factor $1.25$), and reduced post-collision rebound (rest factor $0.9$). ``Treadmill'' environments moved the ground backward at $0.05$ length units per step, which required agents to compensate to maintain forward locomotion. Reduced gravity environments set gravity to $-0.25 \frac{m}{{s}^{2}}$ and damping to $5.0$.

\subsubsection{Scaling internal diversity.}

This section adds detail supporting the results reported in Fig. 4 of the main manuscript.

To create collectives with more than two components, we began by taking the first of the weak connectors trained within uniform two-agent collectives, cloning it three times, and using them to connect five clones of the first of the 100 trained agents. 
Agents were tethered together in a binary tree branching structure: a given agent could have zero, one, or two other agents connected to the left of it. The five-agent collective contains one agent (A) furthest forward. Two more agents are tethered behind it (B and C). The fourth and fifth (D and E) are tethered to the back of B. 
This four-connector, five-agent collective was deployed into each of the 12 novel environments. This process was repeated with the same connector and the second agent, and then repeated again another 98 times. These 100 trials were repeated another 100 times, but now using the second weak connector trained within uniform two-agent collectives. This process was repeated for the third through tenth connector. This yields 10 connectors x 100 collectives = 1000 five-agent collectives. We repeated this entire process to create 1000 10-agents collectives, and then again to create 1000 20-agent collectives. Each data point within the orange dashed lines in Fig. 4 thus represents the mean forward travel of 10 connectors x 100 collectives = 1000 collectives.

This whole process was repeated a second time. But, this time, we began by taking the first of the weak connectors trained within diverse two-agent collectives, cloning it twice, and using the pair to connect together three different randomly-chosen already trained agents. This two-connector, three-agent collective was deployed into each of the 12 novel environments. This process was repeated a second through 100th time for the same connector. Another 100 of these trials were conducted using the second diversity-trained weak connector, and then again for the third connector, until another 10 connectors x 100 collectives = 1000 trials had been conducted. Each data point within the blue solid lines in Fig. 4 thus represents the mean forward travel of 10 connectors x 100 collectives = 1000 collectives. 

\subsection{Supplementary Figures}

\includegraphics[width=\textwidth]{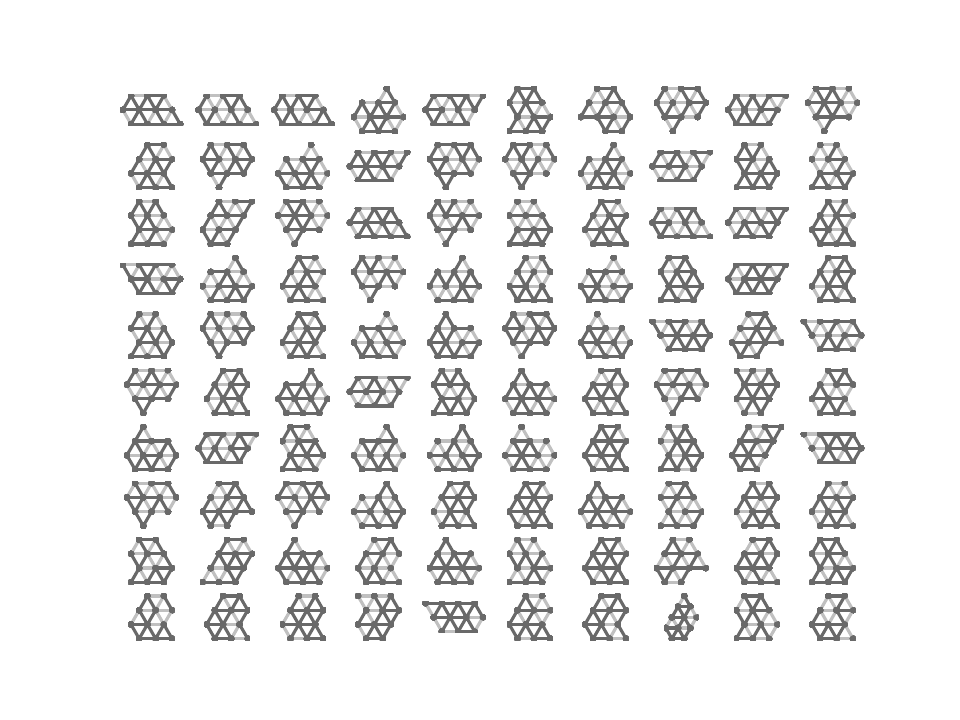}
Supplementary Fig. 1. \textbf{Morphologies of the 100 agents employed for the majority of the experiments.} 
Dark gray lines represent active springs, light gray lines represent passive springs, circles represent mass points.

\includegraphics[width=\textwidth]{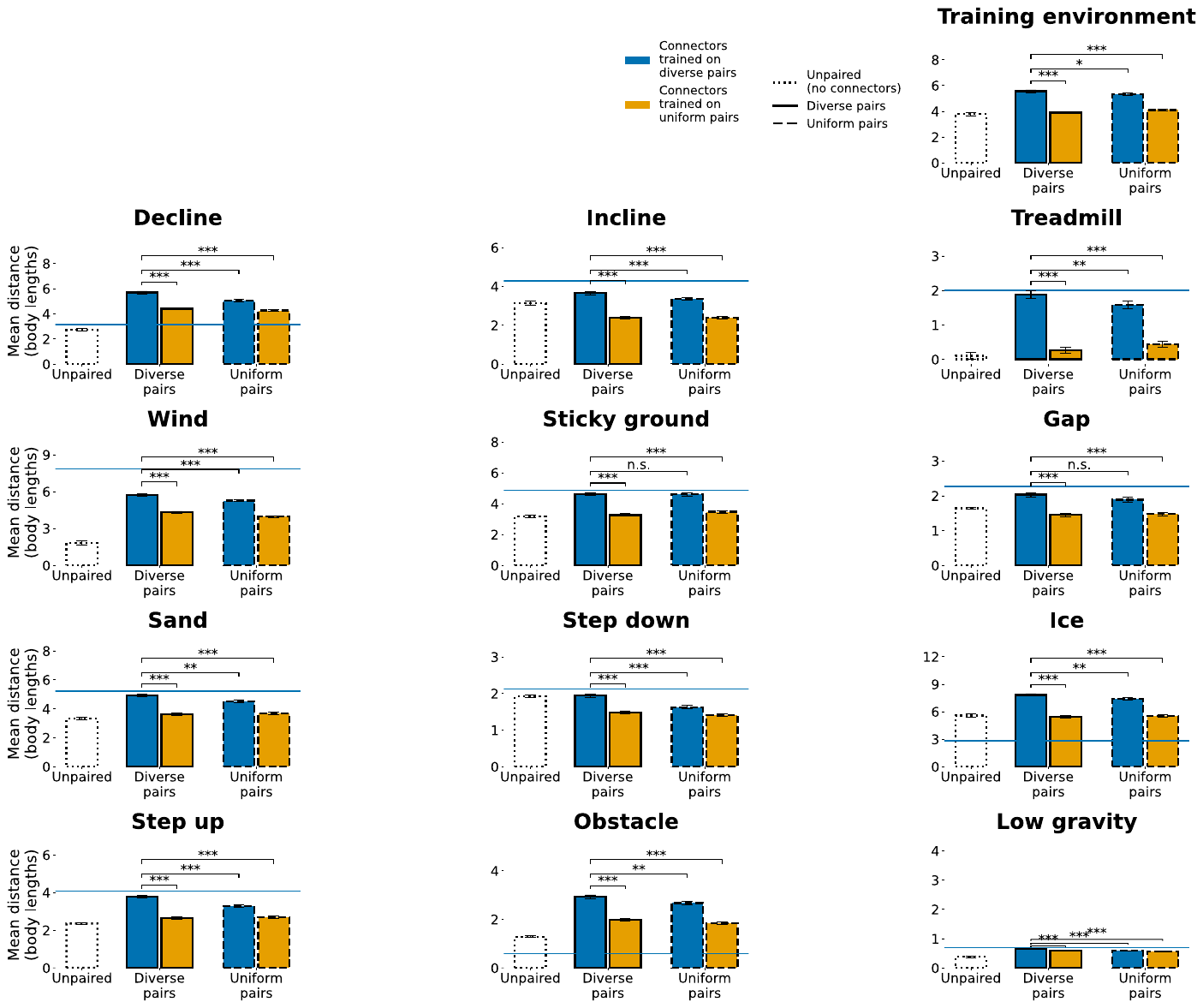}
Supplementary Fig. 2. \textbf{Locomotion performance of collectives in multiple environments}. Each bar represents the mean performance (body lengths traversed) of 10 independently trained connectors, each evaluated on 100 unique agent pairs per environment. The horizontal blue lines indicate ``specialist'' connector performance. This is the average performance of the best among 10 independently trained connectors trained only in each respective novel environment, evaluated in that environment on all 100 agent pairs. This figure extends Figure 2 by illustrating the upper limit achieved by environment-specific specialist training. Error bars denote the standard error of the mean across all trials.

\includegraphics[width=\linewidth]{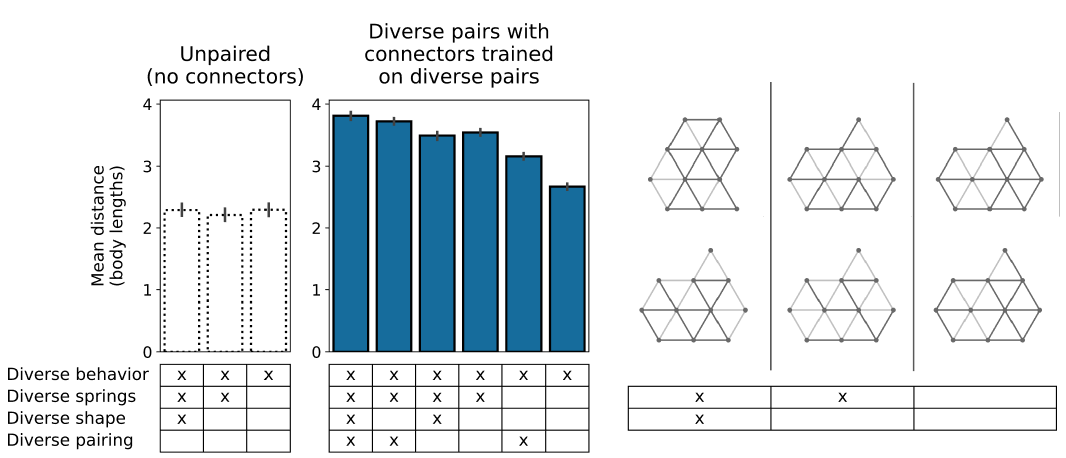}
Supplementary Fig. 3. \textbf{Graded diversity among agents.} Performance of agent pairs connected using diversely trained connectors is evaluated in three diversity settings. Bar plots display mean distance traveled (in body lengths) aggregated across novel environments for unpaired agents (left, white bars, dotted outline) and for connected pairs (right, blue bars, solid outline). Sources of diversity present in each setting are shown in the table below each plot, with ``X'' marks denoting diversity along ``behavior,'' ``springs,'' ``shape,'' and ``pairing.'' Pairing refers to whether agent pairs are made up of clones (uniform) or randomly selected agents (diverse). Images on the right illustrate two example agent morphologies per diversity setting; 100 morphologies per setting were used in the analysis. Results are averaged across 10 independently trained connector models. Error bars indicate bootstrapped 95\% confidence intervals around the mean.

\includegraphics[width=\textwidth]{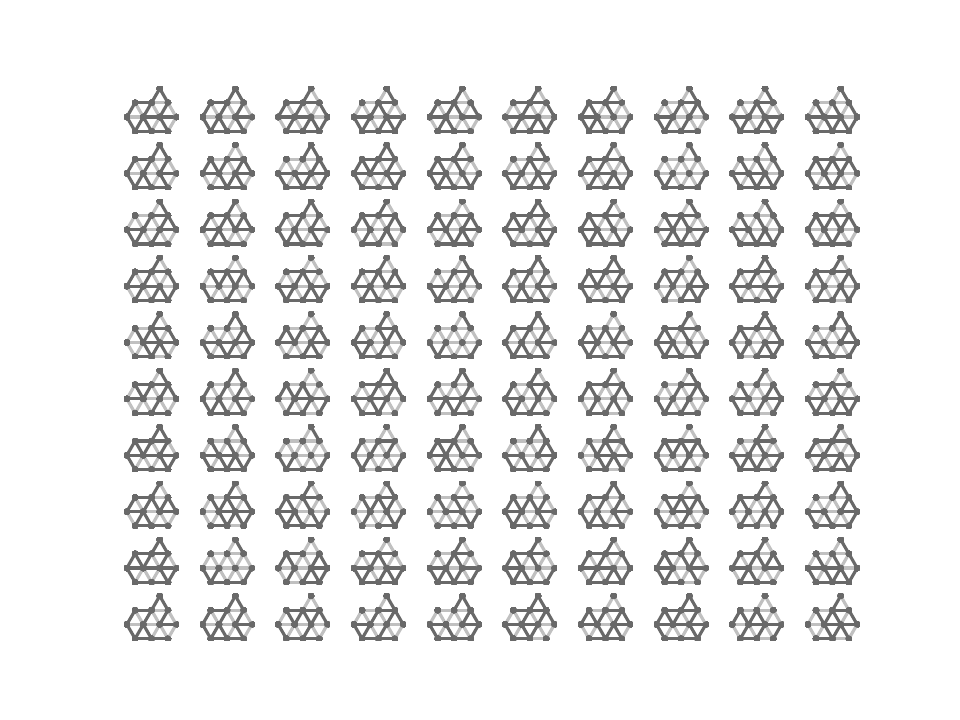}
Supplementary Fig. 4. \textbf{Morphologies of agents with identical shapes but different spring distributions, used for the comparison in Supplementary Fig. 3.} Dark gray lines represent active springs, light gray lines represent passive springs, circles represent point masses.

\includegraphics[width=\textwidth]{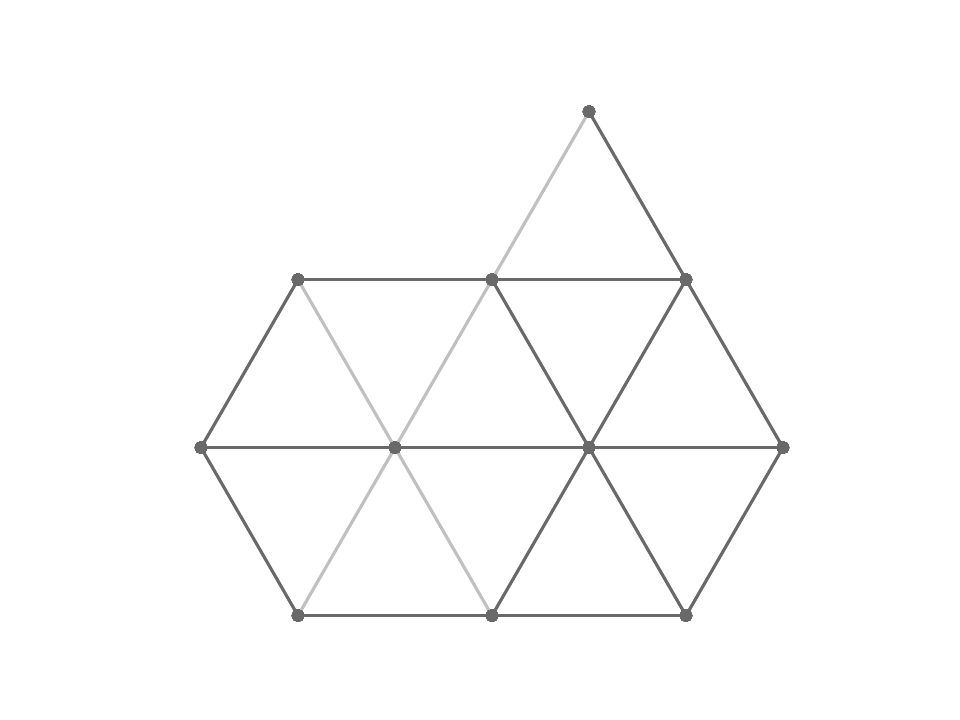}
Supplementary Fig. 5. \textbf{Morphology of agents with identical shapes and identical spring distributions, used for the comparison in Supplementary Fig. 3.} Dark gray lines represent active springs, light gray lines represent passive springs, circles represent point masses.

\newpage
\subsection{Supplementary Table}

``Familiarity'' of hidden states for controller connector models evaluated in unfamiliar environments. For each test environment, the table shows the mean number of 2D states (PaCMAP-reduced embeddings of 32-dimensional connector hidden states, rounded to integer coordinates and aggregated across trials) for 10 independently initialized and trained connector models evaluated on 100 agent pairs per environment. States are classified as: present in both training and novel environments, present only in training, or present only in the novel environment. Columns compare connectors trained on diverse versus uniform agent pairs. For most environments, connectors trained on diverse pairs experience more overlap in dimensionality-reduced state space between training and novel environment (bold, leftmost two columns), while uniformly trained connectors encounter a greater number of novel states exclusive to the novel environment (bold, rightmost two columns).

\begin{tabular}{r||rcl||rcl|rcl}
& \multicolumn{3}{c|}{present in both training} 
& \multicolumn{3}{c|}{present only in training} 
& \multicolumn{3}{c}{present only in novel} \\ 
& \multicolumn{3}{c|}{and novel environment} 
& \multicolumn{3}{c|}{environment} 
& \multicolumn{3}{c}{environment} \\ \hline

                & diverse           &     & uniform    & diverse  &     & uniform          & diverse &     & uniform          \\
Environment     &                   &     &            &          &     &                  &         &     &                 \\
Low gravity     & \textbf{87152.9}  & $>$ & 55331.4    & 49701.8  & $<$ & \textbf{68549.0} & 63012.3 & $<$ & \textbf{75986.6} \\
Gap             & \textbf{159042.8} & $>$ & 151879.1   & 2264.8   & $<$ & \textbf{3556.8}  & 38559.4 & $<$ & \textbf{44431.1} \\
Sticky ground   & 177116.1          &     & 180406.9   & 3385.2   &     & 2542.8           & 19365.7 &     &         16917.3 \\
Ice             & \textbf{178085.9} & $>$ & 176870.6   & 6528.2   &     & 5502.2           & 15252.9 & $<$ & \textbf{17494.2} \\
Obstacle        & \textbf{180642.9} & $>$ & 174706.6   & 1349.6   & $<$ & \textbf{1936.2}  & 17874.5 & $<$ & \textbf{23224.2} \\
Sand            & 170939.1          &     & 173830.6   & 4645.4   &     & 4220.2           & 24282.5 &     &         21816.2 \\
Decline         & \textbf{162845.7} & $>$ & 141239.4   & 3622.8   & $<$ & \textbf{8181.8}  & 33398.5 & $<$ & \textbf{50445.8} \\
Incline         & \textbf{178614.0} & $>$ & 173314.3   & 2698.0   & $<$ & \textbf{6339.6}  & 18555.0 & $<$ & \textbf{20213.1} \\
Step down       & \textbf{140278.1} & $>$ & 135812.3   & 2428.0   & $<$ & \textbf{4565.6}  & 57160.9 & $<$ & \textbf{59489.1} \\
Step up         & \textbf{184086.9} & $>$ & 182431.9   & 224.0    &     & 75.2             & 15556.1 & $<$ & \textbf{17359.9} \\
Treadmill       & 181624.8          &     & 183278.4   & 1794.0   &     & 1181.6           & 16448.2 &     &         15407.0 \\
Wind            & \textbf{158118.0} & $>$ & 131303.1   & 19306.4  & $<$ & \textbf{20748.2} & 22442.6 & $<$ & \textbf{47815.7} \\
\end{tabular}


\subsection{Supplementary Notes}

\paragraph{Caption for Supplementary Movie.}
\textbf{Agent optimization, connector optimization, and collective deployment.}
The first part demonstrates agent optimization.
The second part demonstrates connector optimization.
The third part contrasts the deployment of representative uniform and diverse collectives.
\textbf{\texttt{www.youtube.com/watch?v=Vui-FYUmdoM}}

\end{document}